%% file: main.tex
\pdfoutput=1

\documentclass[11pt]{article}

\usepackage{acl}

\usepackage{times}
\usepackage{latexsym}

\usepackage[T1]{fontenc}

\usepackage[utf8]{inputenc}

\usepackage{microtype}

\usepackage{enumitem}
\usepackage{algorithm}
\usepackage{algpseudocode}
\usepackage{bbm}
\usepackage{amsmath}
\usepackage{multirow}
\usepackage{booktabs}
\usepackage{url}
\usepackage{graphicx}
\definecolor{RoseQuartzBg}{HTML}{F7CAC9}
\definecolor{RoseQuartz}{HTML}{F5A798}
\definecolor{Serenity}{HTML}{92A8D1}
\definecolor{OrangeRed}{rgb}{1.0, 0.27, 0.0}
\definecolor{Red}{rgb}{1.0, 0.0, 0.0}
\definecolor{Turquoise}{HTML}{0F4C81}
\usepackage{xparse}
\usepackage{soul}
\NewDocumentCommand{\lifu}{ mO{} }{\textcolor{OrangeRed}{\textsuperscript{\textit{Lifu}}\textsf{\textbf{\small[#1]}}}}
\NewDocumentCommand{\sijia}{ mO{} }{\textcolor{blue}{{#1}}}
\usepackage{amssymb}
\usepackage{array}
\newcolumntype{L}{>{\centering\arraybackslash}m{4cm}}
\newcolumntype{S}{>{\centering\arraybackslash}m{2cm}}
\newcolumntype{P}{>{\arraybackslash}m{12cm}}
\newcolumntype{Q}{>{\arraybackslash}m{5cm}}
\usepackage{xspace}
\newcommand{\model}{\textbf{\textsc{Talor-EE}}\xspace}
\newcommand{\validation}{\textbf{\textsc{Back-Validation}}\xspace}

\title{Targeted Augmentation for Low-Resource Event Extraction}

\author{Sijia Wang,  \ Lifu Huang
\\
  Virginia Tech
 \\
  {\tt \{sijiawang,lifuh\}@vt.edu}
  }

\begin{document}
\maketitle
\begin{abstract}
Addressing the challenge of low-resource information extraction remains an ongoing issue due to the inherent information scarcity within limited training examples. Existing data augmentation methods, considered potential solutions, struggle to strike a balance between weak augmentation (e.g., synonym augmentation) and drastic augmentation (e.g., conditional generation without proper guidance). This paper introduces a novel paradigm that employs targeted augmentation and back validation to produce augmented examples with enhanced \textit{diversity}, \textit{polarity}, \textit{accuracy}, and \textit{coherence}. Extensive experimental results demonstrate the effectiveness of the proposed paradigm. Furthermore, identified limitations are discussed, shedding light on areas for future improvement\footnote{The source code, model checkpoints, and data are publicly available at \url{https://github.com/VT-NLP/TALOR-EE}.}.

\end{abstract}

\input{sections/1intro}
\input{sections/2relatedWork}

\input{sections/3model}

\input{sections/4experiment}

\input{sections/5conclusion}
\input{sections/limitations}

\section*{Acknowledgements}
This research is supported by the award No. 2238940 from the Faculty Early Career Development Program (CAREER) of the National Science Foundation (NSF). The views and conclusions contained herein are those of the authors and should not be interpreted as necessarily representing the official policies, either expressed or implied, of the U.S. Government. The U.S. Government is authorized to reproduce and distribute reprints for governmental purposes notwithstanding any copyright annotation therein.

\bibliography{ref}
\bibliographystyle{acl_natbib}

\appendix

\input{sections/appendix}


\end{document}

%% file: sections/1intro.tex
\section{Introduction}
\label{sec:intro}

Event extraction (\textbf{EE}) ~\cite{grishman1997information,chinchor1998muc,ahn2006stages} is the task of identifying and categorizing event mentions in natural language text. While supervised methods deliver impressive performance, they depend heavily on extensive manual annotations~\cite{chen-etal-2020-reading, xinyaduEMNLP2020,yinglinACL2020, jianliu2020emnlp, EEMQA_li, Lyu-etal-2021-zero}. Generalizing these approaches to low-resource learning setting poses challenges~\cite{pasupat2014zero, huang2016liberal,huang2020semi,extensively_lai_2020, Adaptive_Shirong_2021, Lyu-etal-2021-zero, zhang-etal-2021-zero, wang-etal-2023-art}.

\begin{figure*}[t]
  \centering
  \includegraphics[width=0.98\textwidth]{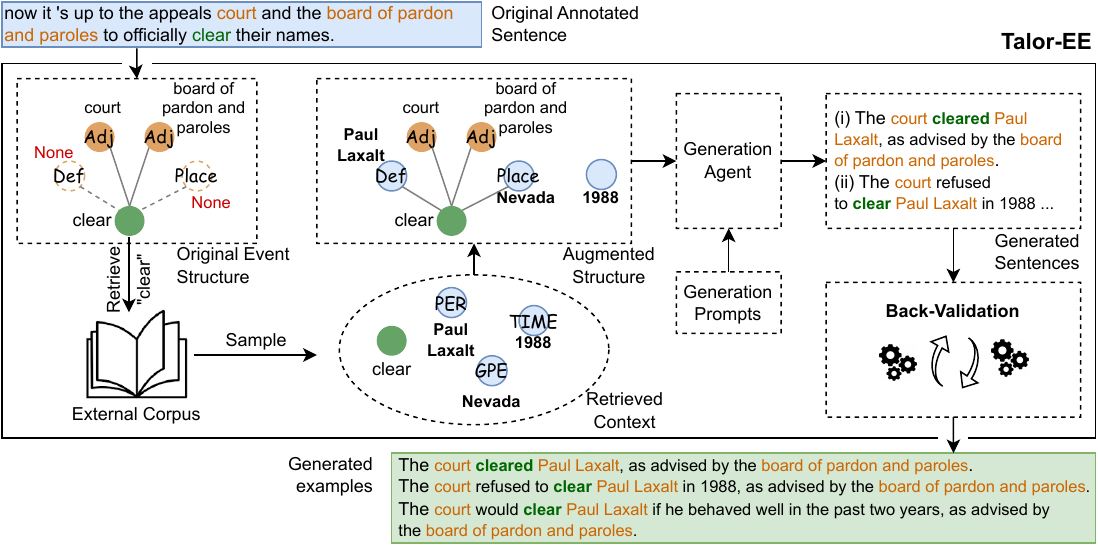}
  \caption{\model framework overview.}
  \label{fig:model}
\end{figure*}

Data augmentation is one direction for efficiently addressing the low-resource event extraction problem. However, it’s remained unexplored what data augmentation strategies are the best for low-resource event extraction given its unique challenges. Previous studies show that weak augmentations, such as synonym augmentation~\citep{ wei-zou-2019-eda} or through back translation~\citep{edunov-etal-2018-understanding}, contribute minimally to distribution enrichment, while drastic augmentations can leading to misguided acquisitions~\cite{ cao-etal-2015-improving, gao2022outofdistribution}. Drastic augmentations usually undermine existing event structure, resulting in grammatical incorrectness, structure misalignment, or semantic drifting~\cite{wang-etal-2023-boosting}.

In this work, we explore several dimensions for data augmentation, including \textit{diversity}, \textit{polarity}, \textit{accuracy}, and \textit{coherence}. 
Our focus revolves around enhancing \textit{diversity} in the context of targeted augmentation for low-resource event extraction (\model). This involves enriching event structures with entities drawn from a targeted subset~\cite{gao2022outofdistribution}. Simultaneously, we address the issue of \textit{polarity} by not only generating positive event mentions based on actual occurrences but also incorporating negative event mentions, e.g., hypothetical event mentions~\cite{ldc_ace05}. This approach is particularly valuable for overcoming limitations in generative event extraction models~\cite{naacl2022degree, liu-etal-2022-dynamic}.
To ensure both \textit{accuracy} and \textit{coherence} in our generated content, we introduce a back-and-forth validation module \validation. The rationale behind this module is that an accurate generation should align with the given event structure, while coherent generation should seamlessly integrate with the same structure.

Our research encompasses a series of comprehensive experiments conducted across various low-resource learning scenarios, including zero-shot and few-shot learning settings. These experiments span different event extraction models. The outcomes of these experiments consistently highlight the effectiveness of targeted augmentation in low-resource event extraction. Notably, among all the dimensions investigated, diversity emerges as the most crucial factor.  Additionally, we meticulously scrutinize the quality of the generated sentences, shedding light on the limitations inherent in the proposed framework.

The contributions of this work are as follows:
\begin{itemize}
\item We explore the application of data augmentation techniques for low-resource event extraction.
\item We develop a novel augmentation method that incorporates enriched event structures and contextual entities, retrieved from external corpus. The generated examples are validated through a back-validation module, ensuring accuracy and coherence.
\item Comprehensive experiments are conducted to assess the effectiveness of the proposed paradigm across various models and datasets.
\end{itemize}

%% file: sections/2relatedWork.tex
\section{Related Work}
\label{sec:related work}
\paragraph{Low-resource Event Extraction} Although some studies have employed meta-learning~\cite{Kang_ICCV_2019, Li2021BeyondMC, Xiao2020FewShotOD, Yan2019MetaRT, Chowdhury2021FewshotIC}, or metric learning~\cite{Sun2021FSCEFO, wang2020few, Zhang2021PNPDetEF, agarwal2021attention} to align candidate event semantics with a few examples of novel event types for few-shot event detection, their performance is inherently constrained by the limited examples provided~\cite{Exploiting_Lai_2020, Meta_Deng_2020, extensively_lai_2020, Cong2021FewShotED, honey_chen_2021, Adaptive_Shirong_2021}.
Recent studies~\cite{wei2023zeroshot, han2023information, li2023evaluating} have explored in-context learning by providing task instructions and a handful of in-context examples. Nevertheless, their experimental findings reveal a notable performance gap between in-context learning and approaches based on fine-tuning.

\paragraph{Data Augmentation} creates synthetic data from the existing data. Traditional data augmentation approaches focus on expanding lexical diversity ~\citep{ wei-zou-2019-eda, feng-etal-2020-genaug,ng-etal-2020-ssmba} or syntax variation \citep{alp_aaai2022, Loem_2022_ExtraPhrase, Hussein_2022_codeswitching, wang-etal-2023-boosting}. Post selection \citep{yang-etal-2020-generative} or representative selection \citep{Edwards_2021_Guiding} helps to prevent a waste of resources and time in generating new documents. Yet existing augmentation methods suffer from gradual drift problem \citep{Hu-etal-2021-semi, Hu-etal-2021-gradient}.  
{The previous work \cite{ma2023star} utilizes language models for training data synthesis but lacks assurance in the soundness and naturalness of event structures due to the random combination of sampled triggers and arguments. Additionally, it falls short by primarily relying on the self-reflection capability of language models, without fully leveraging annotations for existing event annotations.}
Thus, in addition to the lexical and syntactical diversity, we leverage the large-scale pre-trained autoregressive models to generate contextually diversified free texts.

\paragraph{Controlled Text Generation} approaches \citep{ghosh-etal-2021-helpful} generate text with specific constraint. Approaches that promote similarity ~\citep{Guan2021LongTG} or coherence~\citep{Shen2021GTMAG, wang2021adalabel} towards the original sentences lack contextual diversity and might produce over-confident probability estimation ~\citep{wang2021adalabel, gowda-may-2020-finding}. Rule-based constraint generation might generate meaningless tokens to meet constraints \citep{MentionFlags}, while template-based constraint generation  \citep{Cao_2021_acl_controllable} is difficult to generalize to new domains without human effort.

\paragraph{Learning with noisy labels}

Many works learn with noisy labels by detecting corrupted instances, e.g., \cite{Han_coteaching, yu2019does,9008796, Yao2020SearchingTE, 9156369, jiang2020beyond, Zhang2021LearningFN}, and their application to low-resource learning setting~\cite{Wang2020SemiNLLAF, Li2020DivideMix, Cheng2020LearningWI}. However, joint training of the sample selection module and the target task model takes considerable iterations to converge. 
Traditional data-centric methods \cite{Zhu2022DetectingCL} face limitations in low-resource settings due to biased neighbor information. This study demonstrates that training with relatively fair-quality labels can be effective.

%% file: sections/3model.tex
\section{Model}
\label{sec:model}

\subsection{Problem Formulation}
Given a sentence, the Event Extraction (EE) task aims to extract event mentions, represented by an event trigger and a set of event arguments. Formally, given a sentence $w=\{w_1, ..., w_n\}$, and a target event type $e_i$, if there is an event occurrence of $e_i$ in $w$, a EE system aims to extract an event trigger $t$ and its argument mentions $a=\{a_1, ..., a_g\}$.
In this work, we focus on zero-shot and few-shot learning settings of EE. 
For few-shot EE (FSEE), training data contains two parts: (1) A large-scale data set $\mathcal{D}_{base}=\{(\mathbf{x}_i, \mathbf{y}_i)\}_{i=1}^{M}$ that covers the seen event types (named \textit{base types}), where $M$ denotes the number of base event types; (2) a smaller data set $\mathcal{D}_{novel}=\{(\mathbf{x}_j, \mathbf{y}_j)\}_{j=1}^{N\times K}$ that covers $N$ novel event types, with $K$ examples each. Note that the base and novel event types are disjoint except for the \texttt{Other} class, indicating non-event type.
In zero-shot event extraction (ZSEE), the training data set only contains a large-scale set $\mathcal{D}_{base}=\{(\mathbf{x}_i, \mathbf{y}_i)\}_{i=1}^{M}$ for the base event types. The model $f$ will be optimized on base event types and evaluated on the novel types.
Following previous work, we set $N=5, 10$ and $K=0, 1,5,10$ in this work.

\subsection{Targeted Augmentation [Diversity]}
In contrast to previous data augmentation approaches~\citep{ wei-zou-2019-eda, feng-etal-2020-genaug,ng-etal-2020-ssmba, alp_aaai2022, Loem_2022_ExtraPhrase, Hussein_2022_codeswitching, wang-etal-2023-boosting}, we have improved upon the conventional conditional generation method by transitioning from random sampling to a targeted selection strategy. 
{The targeted augmentation module serves as a mechanism to ensure diversity. Theoretically, it can retrieve an infinite number of entities from the external corpus, seamlessly incorporating these entities into the given event structure. Consequently, the module can generate an infinite variety of new event structures. Thus, the targeted augmentation provides a theoretical framework for sampling and augmenting an extensive array of entities, particularly beneficial when working with a limited set of annotated event mentions. }

\paragraph{Dependent Context Retrieval}
For a given event structure, we retrieve context candidates from the corpus that share tokens with the event structure. In our experiments, we gathered sentences containing the mention of the event trigger. To extract context information from the sampled sentences, we utilized the spaCy Named Entity Recognition (NER) parser\footnote{\url{https://spacy.io/usage/linguistic-features}} to identify entity mentions. Consequently, the extracted entity mentions from each sampled sentence serve as context candidates for the given event structure. The context corpus employed in this study is the NYT Annotated Corpus\footnote{\url{https://catalog.ldc.upenn.edu/LDC2008T19}}.

\paragraph{Targeted Generation}

Given an event structure ${e}_i=\{{t}_i, {a}_1, ..., {a}_p\}$ and a sampled context candidate ${c}=\{{c}_1, ..., {c}_q\}$, a generator is leveraged to generate a corresponding sentence. 
If the sampled context entities could potentially serve as argument roles in the original event structures, we employ an add-or-replace strategy, to further tailor the event structure. The feasibility of integrating an entity into the event structure depends on its entity type. If the argument role is vacant in the original structure, and the entity type of the sampled entity aligns with the argument role, we add the entity to the event structure. If the argument role is already populated, we substitute it with the sampled entity.

For example, given an annotation on the sentence \texttt{"now it 's up to the appeals court and the board of pardon and paroles to officially clear their names."}, a \textit{Justice:Pardon} event is represented by the event structure \textit{\{Trigger: clear, Adjudicator: court, Adjudicator: board of pardon and paroles\}}. A complete \textit{Justice:Pardon} structure may also include two argument roles, namely \textit{Defendant} and \textit{Place}. From the sampled context entities \texttt{[Paul Laxalt, 1988, Nevada]}, \texttt{Nevada} is added to the event structure as an \textit{Place} role, and \texttt{Paul Laxalt} is added as a \textit{Defendant} role. Note that \texttt{"Nevada"} is added because it is a GPE entity and a GPE entity is one of the possible entity types for a \textit{Place} role. Similarly, \texttt{Paul Laxalt} is added as a \textit{Defendant} because it is a PER entity.
Here we present a generated sentence with the enriched event structure: \texttt{"The court in Nevada clear Paul Laxalt, as advised by the board of pardon and paroles."} The process is illustrated in Figure \ref{fig:model}.

\begin{figure}[t]
  \centering
  \includegraphics[width=\columnwidth]{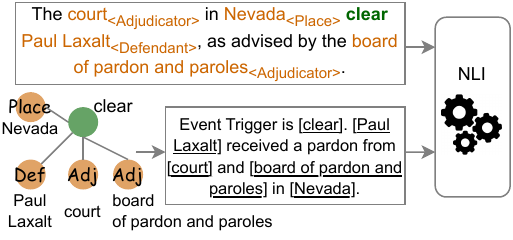}
  \caption{Event mention accuracy verification module.}
  \label{fig:accuracy}
\end{figure}

\subsection{Negative Augmentation [Polarity]}
{Polarity is maintained through the negative augmentation design. This process generates not only positive event mentions but also negative mentions, including hypothetical mentions and believed event mentions. }
For event extraction, we focus on identifying event that occurs, and also negative mentions. For example, in the sentence ``\textit{John Hinkley \textbf{denied} his attempt to \textbf{assassinate} Ronald Reagan.}'', a model, especially generative models, might overlook this Conflict:Attack mention triggered by the token \textit{\textbf{assassinate}}, because this is not an actual event that happens. 
More specifically, negative event mentions include (1) explicit negative mentions: expressed with a negative word such as \textit{not} or \textit{never}, or a negative lexical context such \textit{deny}, \textit{refuse} or \textit{disobey}, (2) asserted mentions: including hypothetical events, believed events, or promised events, etc \cite{ldc_ace05}.

Thus in addition to augmenting high-quality positive training examples, particular attention is paid to augmenting negative training examples. In this work, we write negative/asserted expression prompts to guide their generation. Prompts and generated negative augmentation examples are listed in Table \ref{tab:negative_prompt} and Table \ref{tab:neg_aug_example} in Appendix \ref{appendix:neg_event_mentions}.

\subsection{Back-Validation}
{Given noisy training examples, previous research has utilized methods to detect and rectify corrupted data during training~\cite{Han_coteaching, yu2019does,9008796, Yao2020SearchingTE, 9156369, jiang2020beyond, Zhang2021LearningFN}, but such approaches necessitate extensive training. In our context, where the generated data is considered of reasonable quality, we propose the incorporation of a back-and-forth validation module. This module aims to ensure the \textit{accuracy} and \textit{coherence} of the generated content, thereby enhancing the reliability of the augmented examples.}
\paragraph{Event Mention Accuracy Verification [Accuracy]}
For each generated example, its accuracy can be verified through an entailment verification module. As shown in Figure \ref{fig:accuracy}, given the generated sentence and its source event structure, we first textualize the event structure into a passage to express the event structure, by a pre-defined template~\cite{naacl2022degree}. Then the two texts will be passed into an NLI entailment verification module. The intuition is that, for a valid generation, it should entail the template passage with the event structure.  

\begin{figure}[t]
  \centering
  \includegraphics[width=0.8\columnwidth]{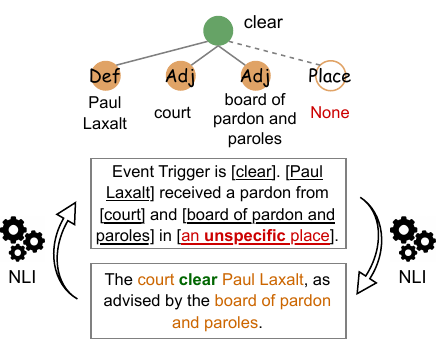}
  \caption{Event mention coherence verification module.}
  \label{fig:coherence}
\end{figure}

\paragraph{Event Mention Coherence Verification [Coherence]}

In addition to ensuring generation accuracy, we aim for the generated sentence to exhibit strong coherence with the provided event structure. Specifically, there should be no extraneous or omitted arguments when compared to the given event structure. The intuition is that if the generated sentence aligns coherently with the provided event structure, a template passage incorporating the event structure should entail the generated sentence, and vice versa.
A distinctive scenario arises when the event structure is incomplete. In such instances, we adapt the missing argument role in the template with the expression "an unspecific [argument role]." Illustrated in Figure \ref{fig:coherence}, if the \textit{Place} argument role is absent, we want to ensure that the generated event mention does not introduce an extraneous arbitrary Place argument role. Consequently, we substitute "[Place]" with "[an unspecific Place]." This modification ensures that the generated sentence fails the forward-and-backward entailment test in such scenarios.

\subsection{Generative Event Extraction Model}
DEGREE~\cite{naacl2022degree} is a generative event extraction model that conceptualizes event extraction as a conditional generation problem. Given a sentence and a crafted prompt, DEGREE generates an output following a specified format. The predictions for event triggers and argument roles can be then parsed from the generated output using a deterministic algorithm. In contrast to earlier classification-based models, the generation framework offers a versatile approach to incorporate supplementary information and guidance. Through the creation of suitable prompts, DEGREE can better capture the dependencies between entities and, consequently diminish the requisite number of training examples.

The EE template defines the anticipated output format and is organized into two main parts. The initial segment is referred to as the trigger template, structured as ``Event trigger is <Trigger>'', with ``<Trigger>'' acting as a placeholder for event trigger in the original passage. The subsequent section is the argument template, and its composition varies based on the specific event type. For instance, the argument template for a Conflict:Attack event is “\underline{some people} or \underline{some organization} in \underline{somewhere} was ordered by \underline{some adjudicator} to pay a fine.” Each underlined string, beginning with "some-," serves as a placeholder corresponding to an argument role for a Justice:Fine event. For example, "somewhere" corresponds to the Place where the event occurs. Note that every event type has its own argument template. 
Event extraction templates and the construction details can be found in \cite{naacl2022degree}.

\subsection{Robust Fine-tuning}
Given the synthesized training samples $\mathcal{D}_{gen}$ that augment $\mathcal D_{train}$ for fine-tuning a classification $M$. 
The primary concern is the presence of label noise, where some generated samples may inaccurately align with their corresponding labels, potentially degrading model performance when using standard supervised learning. 
To address this challenge, we employ a noise-robust training procedure to enhance stability.
We first fine-tune the back-validator $V$ with the training data constructed from the base dataset. For negative examples, we construct two datasets: (1) sample unpaired event structures and sentences within the corpus and (2) replace argument roles in the template with "an unspecific [argument role]". Then we validate the augmented examples with the fine-tuned validator $V$, and validated examples are then used for fine-tuning the EE model $M$. Finally, we employ a random sample selection on the base data set  $\mathcal{D}_{base}$ and the synthesized training set $\mathcal D_{gen}$, along with the entire few shot training set $\mathcal{D}_{novel}$ to update the EE model $M$. The algorithm is shown in Algorithm \ref{alg1}.

\begin{algorithm}[t] 
\small
\caption{Robust Fine-tuning} 
\textbf{Input:} Base data set $\mathcal{D}_{base}$; few shot training set $\mathcal{D}_{novel}$; synthesized training set $\mathcal D_{gen}$.\\
\textbf{Output:} Model $M$, validator $V$

\label{alg1} 
\begin{algorithmic} 
\State fine-tune $V$ with back-validation data constructed from $\mathcal D_{train}$
\State pass $\mathcal D_{gen}$ into $V$, collect $\mathcal D'_{gen}$ that pass back-validation
\For{each epoch $t$}
    \State {Sample meta batch $D^t_{base}$ from $\mathcal{D}_{base}$}
    \State Sample noisy batch $D^t_{gen}$ from $\mathcal D'_{gen}$
    \State Update model $M$ with $D^t_{train}$, $\mathcal{D}_{novel}$, and $D^t_{gen}$
    \State Discard corrupted data by semantic distance to the center instances
\EndFor
\end{algorithmic}
\end{algorithm}

\input{tables/cost}

%% file: tables/cost.tex
\begin{table}[t]
\begin{center}
\small
\begin{tabular}{lccc}
\toprule
Model & Time/Sentence(s) & Cost/Sentence(\$)
\\\midrule

Vicuna-7B & 2.7 & 0\\\midrule
LLaMA2-7B & 8.7 &0\\\midrule
GPT-3.5-turbo& 2.4& $\sim$0.0035\\

\bottomrule
\end{tabular}

\caption{Augmentation cost per sentence.}
\label{tab:cost}
\end{center}
\end{table}

%% file: sections/4experiment.tex
\input{tables/few_shot_ace}

\section{Experiments}
\label{sec:experiments}

\begin{figure*}[t]
  \centering
  \includegraphics[width=\textwidth]{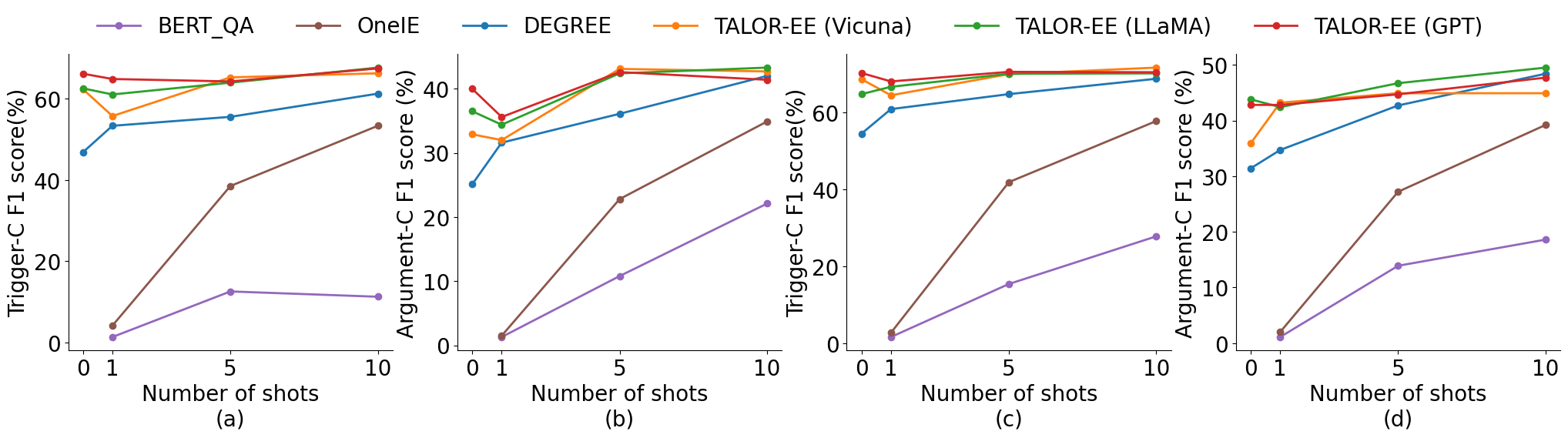}
  \caption{Experimental results on ACE05-E. (a-b) are visualizations for Common 5, and (c-d) for Common 10.}
  \label{fig:ace}
\end{figure*}

\input{tables/few_shot_ere}

We perform experiments on three public benchmark datasets, including ACE05-E (Automatic Content Extraction)\footnote{\url{https://catalog.ldc.upenn.edu/LDC2006T06}} and ERE (Entity Relation Event)~\cite{song2015light}. To showcase the effectiveness of the proposed method under low resource settings, experiments are conducted under $N$way-$K$shot learning setting, where $N\in\{5, 10\}$, and $K\in \{0,1,5,10\}$.

\paragraph{Compared baselines} We consider the following baselines: (1) Matching baseline\footnote{(1) and (2) are baselines for event detection tasks, thus only trigger detection results are reported.}, a proposed baseline that makes trigger predictions by performing string matching between the input passage and the event keywords. (2) Lemmatization baseline, another proposed baseline that performs string matching on lemmatized input passage and the event keywords. (3) BERT\_QA\cite{xinyaduEMNLP2020}, (4) OneIE~\cite{yinglinACL2020}, (5) DEGREE~\cite{naacl2022degree} and (6) QueryExtract \cite{WangAcl2022_query}. The implementation details can be found in Appendix \ref{sec:Implementation}.

\paragraph{Generation Agents} Three generation agents are experimented in this work, including \texttt{vicuna-7b-v1.3} (Vicuna), \texttt{Llama-2-7b} (LLaMA), and \texttt{gpt-3.5-turbo} (GPT).
For each agent, we list the augmentation cost in Table \ref{tab:cost}, where two factors are listed including generation time and cost per sentence.

\begin{figure*}[t]
  \centering
  \includegraphics[width=\textwidth]{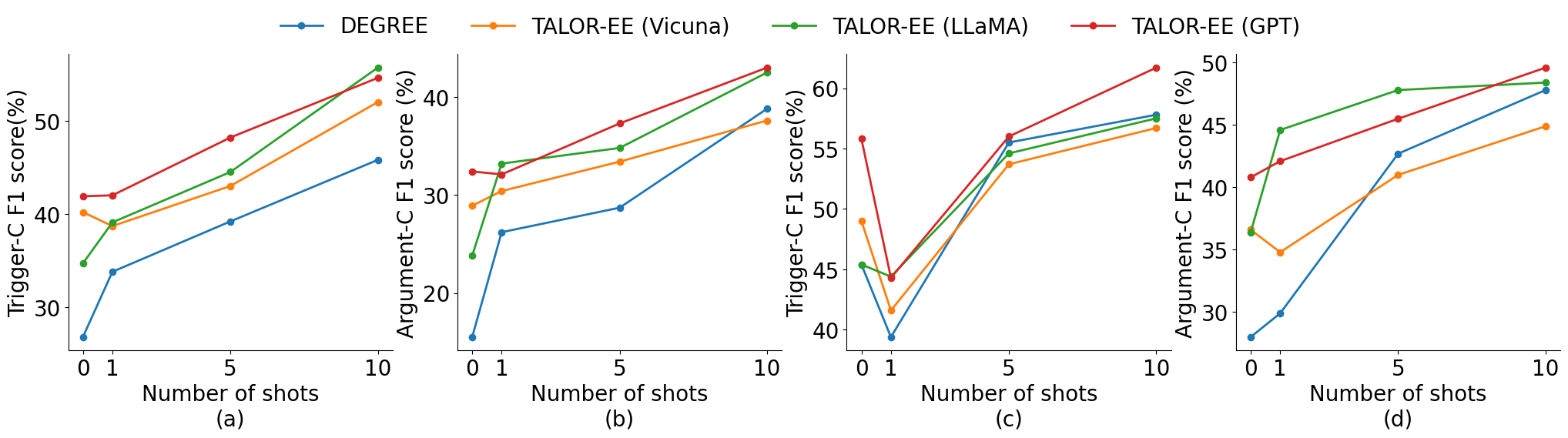}
  \caption{Experimental results on ERE. (a-b) are visualizations for Common 5, and (c-d) for Common 10.}
  \label{fig:ere}
\end{figure*}

\input{tables/ablations}

\subsection{Main results}

The experimental results for low-resource Event Extraction (EE) are presented in Table \ref{tab:fs_ace} and Figure \ref{fig:ace} for ACE05-E,  and Table \ref{tab:fs_ere} and Figure \ref{fig:ere}  for ERE, respectively. From the experiment results, several conclusions can be drawn:
(1) With the augmented examples, the performance of low-resource EE generally exhibits improvement, evident in both zero-shot learning and few-shot learning settings. This improvement is consistent across different generation agents (Vicuna, LLaMA, and GPT) and backbone EE models. Table \ref{tab:fs_ace_qe} displays experimental results on ACE05-E with QueryExtract as the backbone model, highlighting the effectiveness of augmented training examples across various EE models.
(2) The observed improvement is more pronounced in extremely low-resource scenarios, particularly in zero-shot, 1-shot, and 5-shot scenarios. The impact is less significant when more clean training examples are available, such as in the 10-shot setting.
(3) We observe that the performance of zero-shot augmented training can surpass that of 1-shot training with clean examples. 
This discrepancy arises because some sampled clean training examples may not straightforwardly express event information. For instance, the token ``open'' could trigger a \textit{Start-Organization} event, introducing confusion in the semantics of the \textit{Start-Organization} event type.
(4) Augmented examples generated by different generation agents consistently enhance low-resource EE performance. Notably, greater performance gains are achieved with examples generated by LLaMA and GPT.

{Additionally, we have evaluated the generation quality and the effectiveness of the proposed modules. Notably, for diversity, there is a substantial increase in unique argument roles compared to the few-shot examples. For example, in the common 10 and 5-shot settings, the count of unique argument roles surged from 142 to 1184, marking a remarkable increase of 2502 percentage points, on average across the generation models. Regarding polarity, among the 30 sampled augmentations verified through human evaluation, the generated event mention expressions consistently align with the targeted negative expression types. In terms of back-validation, the evaluation involved two annotators who each assessed 200 randomly sampled generations (100 for with back-validation generations and 100 for generations without back-validation). On average, seven generations were deemed not fluent when utilizing the back-validation module, while 19 generations were identified as not fluent without the back-validation module.}

\input{tables/error}

\subsection{Ablation Studies}

An ablation study was conducted to assess the effectiveness of each proposed module, and the experimental results are presented in Table \ref{tab:ablations_ace}. {(Omitting the enriched context in the setting entails bypassing the Dependent Context Retrieval module, resulting in the absence of newly generated event structures.)}
On average, across all settings, the performance of trigger classification decreased by 2.5\% and 1.9\%, and argument classification decreased by 8.3\% and 7.1\%, in the absence of enriched context or back-validation, respectively. 
Without negative augmentations, the argument classification decreases by 7.5\%, while trigger classification performance is on par with \model (LLaMA). This highlights that the designed modules have a more pronounced impact on argumentation classification than on trigger detection. The absence of enriched context led to the most significant decrease in argument classification performance, emphasizing the crucial role of augmentation diversity in mitigating low-resource argument extraction.

\subsection{Error Analysis}

Table \ref{tab:error} illustrates several challenging examples. 
For event trigger detection, most of the errors are from the insufficient understanding of the trigger phrase. For example in example (a) in Table \ref{tab:error}, linking the phrase ``crumbling'' to the \textit{End-Org(anization)} event is challenging given the limited trigger training examples from either clean data or augmented data. Example (b) is challenging because the token ``combination'' entails closer semantic relation to the \textit{Merge-Org} event. {Example (c) illustrates a case where the current data augmentation model falls short in generating intricate event expressions. Example (d) illustrates a scenario in which the use of augmented data could potentially cause confusion. In this case, the actual event pertains to a film release rather than a judicial release. Despite inadequate context information, there is a likelihood that the augmented data might have generated a false prediction with increased confidence. One potential solution to this challenge is the ability to distinguish between multiple meanings of the same word. }

In contrast to event trigger detection, argument extraction presents greater challenges, as improvements in argument extraction prove less pronounced than those in trigger detection. Our conclusion stems from a meticulous analysis of the generated outputs and prediction results, revealing two primary reasons. 
The first reason is the lack of clear and comprehensive explanations for certain argument roles, for example, the argument role ``agent'' in the \textit{Start-Org} event type.  According to the definition \cite{ldc_ace05}, an ``agent'' in a \textit{Start-Org} event is a ``PER'', ``ORG'', or ``GPE'' entity responsible for the ``START-ORG'' Event. However, it requires tremendous expert knowledge to write precise instructions for argument roles like this.
The second reason pertains to the lack of clear distinctions among argument roles in generation prompts. We recognize that elucidating the purpose and differentiation of each argument role can be intricate. For instance, we observed minimal or even adverse effects of augmented data on the event type ``Transfer-Ownership''. This complexity arises from the potential confusion surrounding three specific argument roles: ``Beneficiary'', ``Buyer'', and ``Seller'', particularly when the trigger involves terms like ``sell'' or ``acquire''. Notably, altering the trigger from ``sell'' to ``acquire'' induces a substantial change in the sentence's entire syntactic structure.

%% file: tables/few_shot_ace.tex
\begin{table*}[ht]
\centering
\small
\begin{tabular}{l|c|c c|c c|c c|cc}
\toprule
    {\multirow{2}{*}{\textbf{Method}}}&\multirow{2}{*}{\textbf{K-shot}} & \multicolumn{4}{c|}{Common 5}&\multicolumn{4}{c}{Common 10}\\
    \cmidrule{3-6}\cmidrule{7-10}
    &&Tri-I &  Tri-C& Arg-I& Arg-C& Tri-I&  Tri-C & Arg-I  & Arg-C \\
\midrule
    Matching Baseline       &full& 42.7 & 42.1  & -  & -    & 46.3   &46.3 & -  &-\\
    Lemmatization Baseline  &full& 51.5 & 50.2  & -  & -    & 56.0   &56.0 & -  &-\\
    OneIE                   &full& 72.7   & 70.5  & 52.3  & 49.9  & 74.5   &73.0 & 51.2  & 48.9 \\
    DEGREE                  &full& 68.4   & 66.0  & 51.9  & 48.7  & 72.0   &69.8 & 52.5  & 49.2 \\
\midrule
    \multirow{4}{*}{BERT\_QA}
    & 1-shot     & 10.0   & 1.4   & 1.3   & 1.3   & 8.2   & 1.6  & 1.1   & 1.1   \\
    & 5-shot     & 14.0   & 12.6  & 11.1  & 10.8  & 20.8  & 15.4 & 14.6  & 13.9  \\
    & 10-shot    & 37.8   & 11.3  & 22.9  & 22.1  & 32.0  & 27.8 & 19.5  & 18.6  \\
\midrule
    \multirow{4}{*}{OneIE}
    & 1-shot        & 4.2    & 4.2   & 1.5   & 1.5   & 4.1   & 2.7  & 2.0   & 2.0   \\
    & 5-shot        & 39.3   & 38.5  & 24.8  & 22.8  & 41.9  & 41.9 & 29.7  & 27.2  \\
    & 10-shot       & 54.8   & 53.3  & 36.0  & 34.9  & 61.5  & 57.8 & 41.4  & 39.2  \\
\midrule
    \multirow{4}{*}{DEGREE}
    & 0-shot       & 53.3   & 46.8  & 29.6  & 25.1  & 60.9  & 54.5 & 42.0  & 31.4  \\
    & 1-shot       & 60.1   & 53.3  & 38.8  & 31.6  & 61.2  & 60.9 & 41.1  & 34.7  \\
    & 5-shot       & 57.8   & 55.5  & 40.6  & 36.1  & 65.8  & 64.8 & 45.3  & 42.7  \\
    & 10-shot      & 63.8   & 61.2  & 46.0  & 42.0  & 72.1  & 68.8 & \textbf{52.5}  & 48.4  \\
\midrule
    \multirow{4}{*}{\model(Vicuna)}
    &0-shot &  66.1  & 62.3  & 38.7  & 32.9  & 71.6  & 68.7  & 40.7  & 35.9  \\
    &1-shot & 63.5   & 55.7  & 37.5  & 32.0  & 69.2  & 64.5  & \textbf{47.8}  & \textbf{43.2}  \\
    &5-shot & 67.0   & \textbf{65.2}  & \textbf{46.6}  & \textbf{43.1}  & 72.7  & 70.0  & 50.1  & 44.9  \\
    &10-shot& \textbf{70.4}   & 66.2  & \textbf{46.4}  & 42.7  & 73.9  & \textbf{71.7}  & 49.2  & 44.9  \\
\midrule
    \multirow{4}{*}{\model(LLaMA)}
    &0-shot & 65.0   & 62.5  & 41.0  & 36.5  & 65.6  & 64.8  & \textbf{47.5 } & \textbf{43.8}   \\
    &1-shot & 66.5   & 61.0  & \textbf{42.3}  & 34.4  & 71.5  & 66.7  & 45.4  & 42.4   \\
    &5-shot & \textbf{70.2 }  & 63.9  & 46.3  & 42.4  & 71.7  & 70.1  & \textbf{50.5}  & \textbf{46.7 }  \\
    &10-shot& 70.0   & \textbf{67.6}  & 46.2  & \textbf{43.3}  & 70.5  & 70.2  & 51.2  & \textbf{49.5 }  \\
\midrule
    \multirow{4}{*}{\model(GPT) }
    &0shot  & \textbf{67.9}   & \textbf{66.1}  & \textbf{46.1}  & \textbf{40.0}  & \textbf{72.5}  & \textbf{70.3}  & 46.9  & 42.8  \\
    &1-shot & \textbf{68.5}   & \textbf{64.8}  & 42.1  & \textbf{35.6}  & \textbf{72.5}  & \textbf{68.1}  & 46.5  & 42.8  \\
    &5-shot & 67.9   & 64.2  & 44.6  & 42.6  & \textbf{73.6}  & \textbf{70.6}  & 48.5  & 44.7  \\
    &10-shot& 70.2   & 67.4  & 43.0  & 41.4  & \textbf{74.2}  & 70.5  & 48.3  & 47.7  \\
\bottomrule
\end{tabular}
\caption{Low-resource EE results on ACE05-E. Bold represents the highest score for the current setting. }
\label{tab:fs_ace}
\end{table*}

%% file: tables/few_shot_ere.tex
\begin{table*}[ht]
\centering
\small
\begin{tabular}{l|c|c c|c c|c c|cc}
\toprule
    \multirow{2}{*}{\textbf{Method}} &\multirow{2}{*}{\textbf{K-shot}}& \multicolumn{4}{c|}{Common 5}&\multicolumn{4}{c}{Common 10}\\
    \cmidrule{3-6}  \cmidrule{7-10}
    &&Tri-I &  Tri-C& Arg-I& Arg-C& Tri-I&  Tri-C & Arg-I  & Arg-C \\
\midrule
    DEGREE &full          & 54.7   & 53.1  & 45.4  &  44.7  & 58.8  & 58.2  & 51.3  & 50.8 \\
\midrule
\multirow{4}{*}{DEGREE}
    & 0-shot  & 32.2   & 26.8  & 16.1  & 15.5  & 47.7  & 45.4  & 28.7  & 28.0 \\
    & 1-shot  & 34.4   & 33.8  & 28.0  & 26.2  & 39.4  & 39.4  & 30.7  & 29.9  \\
    & 5-shot  & 44.8   & 39.2  & 28.9  & 28.7  & 56.3  & 55.5  & 44.5  & 42.7  \\
    & 10-shot & 48.4   & 45.8  & 39.3  & 38.8  & 59.3  & 57.8  & 48.4  & 47.8  \\
\midrule
\multirow{4}{*}{\model(Vicuna)}
    &0-shot  & 41.9  & 40.2  & 31.0  & 28.9  & 50.6  & 49.0  & 37.9  & 36.6  \\
    &1-shot  & 48.5   & 38.7  & 31.3  & 30.4  & 47.8   & 41.6  & 35.9  & 34.8  \\
    &5-shot  & 45.8   & 43.0  & 35.8  & 33.4  & 56.2  & 53.7  & 42.5  & 41.0  \\
    &10-shot & 55.7  &  52.0 &  40.6 & 37.6  & 58.2  & 56.7  & 47.8  & 44.9  \\
\midrule
\multirow{4}{*}{\model(LLaMA)}
    &0-shot  & 40.8  & 34.7  & 26.2  & 23.8  & 51.6  & 45.4  & 37.8  & 36.4  \\
    &1-shot  & 47.4  & 39.1  & 33.4  & \textbf{33.2}  & 47.3  & \textbf{44.4}  & \textbf{46.2}  & \textbf{44.6}  \\
    &5-shot  & 48.9  & 44.5  & 37.7  & 34.8  & 55.3  & 54.6  & 48.5  & \textbf{47.8}  \\
    &10-shot & \textbf{58.1}  & \textbf{55.7}  & \textbf{45.5}  & 42.5  & 58.2  & 57.5  & 52.2  & 48.4  \\
\midrule
\multirow{4}{*}{\model(GPT)}
    &0-shot  & \textbf{49.3}   & \textbf{41.9}  & \textbf{34.0}  & \textbf{32.4}  & \textbf{57.1}   & \textbf{55.8}  & \textbf{43.1}  & \textbf{40.8 } \\
    &1-shot  & \textbf{50.3}   & \textbf{42.0}  & \textbf{34.5}  & 32.1  & \textbf{51.6}  & 44.3  & 43.7  & 42.1  \\
    &5-shot  & \textbf{52.9}   & \textbf{48.2}  & \textbf{39.1}  & \textbf{37.3}  & \textbf{57.5}  & \textbf{56.0}  & \textbf{49.4}  & 45.5  \\
    &10-shot & 56.9   & 54.6  & 43.5  & \textbf{43.0 } & \textbf{62.4}  & \textbf{61.7}  & \textbf{53.4}  & \textbf{49.6}  \\
\bottomrule
\end{tabular}
\caption{Low-resource EE results on ERE. Bold represents the highest score for the current setting. }
\label{tab:fs_ere}
\end{table*}

%% file: tables/ablations.tex
\begin{table*}[ht]
\centering
\small
\begin{tabular}{l|c|c c|c c|c c|cc}
\toprule
    \multirow{3}{*}{\textbf{Method}}& \multirow{3}{*}{\textbf{K-shot}} & \multicolumn{4}{c|}{Common 5}&\multicolumn{4}{c}{Common 10}\\
    \cmidrule{3-6}  \cmidrule{7-10}
    &&Tri-I &  Tri-C& Arg-I& Arg-C& Tri-I&  Tri-C & Arg-I  & Arg-C \\
\midrule
    \multirow{3}{*}{\model(LLaMA)}
    &{ 1-shot}  & 66.5   & 61.0  & 42.3  & 34.4  & 71.5  & 66.7  & 45.4  & 42.4   \\
    &{ 5-shot}  & 70.2   & 63.9  & 46.3  & 42.4  & 71.7  & 70.1  & 50.5  & 46.7   \\
    &{ 10-shot} & 70.0   & 67.6  & 46.2  & 43.3  & 70.5  & 70.2  & 51.2  & 49.5   \\
\midrule
    \multirow{3}{*}{{\;\;- enriched context} }
    &{ 1-shot}  & 61.2   & 52.1  & 35.9  & 28.3  & 72.9  & 64.6  & 46.2  &  40.6   \\
    &{ 5-shot}  & 68.5   & 64.2  & 43.5  & 41.1  & 73.2  & 70.0  & 45.7  &  44.6   \\
    &{ 10-shot} & 67.0   & 63.4  & 43.1  & 39.5  & 74.7  & 71.7  & 46.4  &  43.2   \\
\midrule
    \multirow{3}{*}{\;\;- negative augmentations}  
    &{ 1-shot}   &70.5  & 65.1  & 41.8   & 34.4  & 74.1  & 67.4  & 44.4  & 38.8  \\
    &{ 5-shot}   &69.3  & 62.6  & 41.8   & 39.3  & 77.4  & 73.4  & 48.4  & 42.8  \\
    &{ 10-shot}  &69.1  & 61.3  & 40.8   & 39.6  & 74.1  & 70.5  & 46.6  & 44.3   \\
\midrule
    \multirow{3}{*}{{\;\;- back-validation} }
    &{ 1-shot}  & 61.2   & 52.1  & 35.9  & 28.3  & 72.7  & 66.0  & 47.3  & 42.2    \\
    &{ 5-shot}  & 68.0   & 62.8  & 43.1  & 38.6  & 76.1  & 74.6  & 48.6  & 44.4    \\
    &{ 10-shot} & 67.2   & 65.2  & 42.1  & 40.2  & 75.3  & 71.2  & 47.3  & 46.7    \\
\bottomrule
\end{tabular}
\caption{Ablation study on ACE05-E.}
\label{tab:ablations_ace}
\end{table*}

%% file: tables/error.tex
\begin{table*}[ht!]
\begin{center}
\small
\begin{tabular}{cQLL}
\toprule
\textbf{ID} & Text & GTH & Predictions \\
\midrule

(a) & Hoon said Saddam 's regime was \textbf{crumbling} under the pressure of a huge air assault . & crumbling; End-Org; regime: Org; & None 
\\\midrule
(b) & The \textbf{combination} of the banking operations of Barclays Spain and Zaragozano will bring together two complementary businesses. 
& combination; Transfer-Ownership; Barclays Spain: Buyer; Zaragozano: Artifact;
& combination, Merge-Org; businesses, Org

\\\midrule
(c) 
& \textbf{Married} for the second time , Hariri has five children. 
& Married, Marry; 
Hariri: Person; & None
\\\midrule
(d)&However the firm announced on Friday that it had reached a deal with the British arm of French distributors Pathe to show four \textbf{releases}. & None &
releases; Release-Parole; 
firm: Entity;\\

\bottomrule
\end{tabular}

\caption{Case study for challenging examples}
\label{tab:error}
\end{center}
\end{table*}

%% file: sections/5conclusion.tex
\section{Conclusion}
\label{sec:conclusion}

In conclusion, this study proposes a new paradigm for tackling low-resource event extraction tasks. 
Generation agents are employed to create a diverse training dataset for event structures enriched with domain-invariant entities. The generated examples undergo a thorough back-and-forth validation process to assess accuracy and coherence. Our research encompasses extensive experiments in diverse low-resource learning scenarios, such as zero-shot and few-shot learning settings, across various event extraction models. The outcomes of these experiments highlight the effectiveness of the proposed framework. Furthermore, our proposed methodology can inspire researchers from diverse domains to embrace a comparable paradigm or delve into the investigation of data augmentation methods as a means of enriching their training datasets. 

%% file: sections/limitations.tex
\section*{Limitations}

\model establishes a powerful starting point for advancing few-shot learning research, offering a flexible framework for framing new tasks through our proposed augmentation method. It encourages a systematic exploration of general and resilient enhancements for low-resource event extraction systems. However, augmenting non-event examples takes appropriate attention, as the proposed system may tend to predict additional event mentions. The absence of a clear distinction between an actual event and a non-event mention, due to the lack of a precise definition, underscores the need for appropriate action.
We extend a warm invitation to future low-resource research endeavors and augmentation methods to delve into the structural aspects of event generation within a contrastive setting.

%% file: sections/appendix.tex
\newpage

\begin{table*}[ht]
\centering
\small
\begin{tabular}{l|P}
\toprule
    Event Expression Type & Instruction Prompt\\\midrule
    Negative Events & An Event is NEGATIVE when it is explicitly indicated that the Event did not occur. Negative example 1: His wife was sitting in the backseat and was 'not hurt'. Negative example 2: Yeltsin ordered Skuratov's suspension, but parliament repeatedly 'refused to sack' him. Given the generated sentence, ``[SENT]'', change it into a negative expression that the Event did not occur. \\\midrule
    Believed Events & Believed Events are event mentions that some people or organizations think or believe would happen but are not necessarily real or true event occurrences. Example 1: Rumors of 'arrests' circulated in Vancouver. Example 2: The charity was suspected of 'giving' money to al Qaeda. Given the generated sentence you provide, '[SENT]', change it into a believed event sentence: \\\midrule
    Hypothetical Events & Hypothetical events are event mentions that are supposed to happen but are not necessarily real or true event occurrences. Example 1: Should he not 'pay' the money, they would 'kill' him. Example 2: A demonstration of how he would behave if he were to 'become' President. Given the generated sentence you provide, '[SENT]', change it into a hypothetical event sentence: \\\midrule
    Promised Events & Promised Events are event mentions that are promised to happen but are not necessarily real or true event occurrences. Example 1: He said he would 'leave' town. Example 2: Promises of 'aid' made by Arab and European countries. Given the generated sentence you provide, '[SENT]', change it into a promised event sentence: \\\midrule
    Desired Event & Desired events are event mentions that are desired to happen but not necessarily real or true event occurrences. Example: They wanted to 'acquire' the company last year. Given the generated sentence you provide, ``[SENT]'', change it into a Desired event sentence: \\
\bottomrule
\end{tabular}
\caption{Negative/asserted expression generation template. ``[SENT]'' is a placeholder for the generated sentence with a positive event expression. The instruction is adapted from \cite{ldc_ace05}.}
\label{tab:negative_prompt}
\end{table*}

\input{tables/neg_aug_example}

\section{Implementation} 
\label{sec:Implementation}
For a fair comparison with baseline approaches, we use the pre-trained \texttt{bert-large-uncased} model for fine-tuning and optimizing our model with BertAdam. We optimize the parameters with grid search: training epoch 10, learning rate $\in [3e\text{-}6, 1e\text{-}4]$, training batch size $\in\{8, 12, 16, 24, 32\}$, dropout rate $\in\{0.4, 0.5, 0.6\}$. Our experiments run on one Quadro RTX 8000. For trigger detection, the average runtime is 3.0 hours. For argument detection, the average runtime is 1.3 hours. We use Spacy to generate POS tags. We use three random seed 0, 39, 42 for all experiments, and report the mean scores.
\paragraph{Sampling Strategy}

Note that in the context of few-shot learning with an $N$way-$K$shot setting, the variable K denotes the number of event mentions rather than training examples. The original corpus contains numerous instances where a single sentence includes multiple event mentions, presenting a challenge for the few-shot example sampling process. Without regularization, the sampled examples may probably exceed the specified K event mentions.

To address this issue and ensure that, for every setting, the sampled examples with novel event types do not surpass K, we employ a sorting mechanism based on the frequency of event types in decreasing order. This involves sorting the event types and then sampling in the sorted order.
For instance, consider the examples with "Justice:Acquit" mentions, one of which also includes a "Justice:Convict" mention. If we were to first sample examples for "Justice:Convict" and this particular example is omitted, we would miss the opportunity to include this crucial instance for "Justice:Acquit." This becomes especially significant in settings such as 5-shot or 10-shot, where "Justice:Acquit" has a total of four examples. Without this sampling approach, the mentioned example may be excluded from the training procedure, impacting the model's performance.

\paragraph{Generation Instruction}
The following instruction are used to prompt generations given the event structure: ``You are a helpful assistant in generating fluent and reasonable sentences with event mentions. An Event is a specific occurrence involving participants. An Event is something that happens. An Event can frequently be described as a change of state. Please be sure the given event information is in the generated sentence. However, the given context information is optional in generation. Generate a sentence with \{event\_type\_name\} event, with optional context information: \{list\_of\_context\_entitites\}. \{event\_template\}.'' The \{event\_template\} refers to the textual representation given the event structure, as presented in \cite{naacl2022degree}.

\begin{table*}[ht]
\centering
\small
\begin{tabular}{l|c|c c|c c|c c|cc}
\toprule
    \multirow{2}{*}{\textbf{Method}} &\multirow{2}{*}{\textbf{K-shot}} & \multicolumn{4}{c|}{Common 5}&\multicolumn{4}{c}{Common 10}\\
    \cmidrule{3-6}  \cmidrule{7-10}
    &&Tri-I &  Tri-C& Arg-I& Arg-C& Tri-I&  Tri-C & Arg-I  & Arg-C \\
\midrule
    \multirow{3}{*}{QE }
    & 1-shot  & 58.6   & 48.7  & 33.1  & 29.3  & 58.6  & 51.2  & 37.5  & 30.1\\
    & 5-shot  & 61.9   & 57.1  & 37.6  & 33.1  & 66.7  & 61.1  & 41.7  & 36.5\\
    & 10-shot & 64.1   & 62.2  & 40.3  & 38.6  & 72.0  & 67.2  & 45.6  & 45.2 \\
\midrule
   \multirow{3}{*}{\textbf{\textsc{Tolar-QE}} (Vicuna) }
    & 1-shot  & 60.6   & 58.0  & 41.8  & 34.2  & 60.4  & 58.0  & 41.4  & 35.0 \\
    & 5-shot  & 65.4   & 62.1  & 44.3  & 35.8  & 70.8  & 68.8  & 47.2  & 41.6 \\
    & 10-shot & 65.7   & 64.0  & 43.4  & 39.6  & 69.5  & 68.1  & 50.8  & 43.7 \\
\midrule
   \multirow{3}{*}{\textbf{\textsc{Tolar-QE}} (LLaMa) }
    & 1-shot  & 64.7   & 57.6  & 39.3  & 28.3  & 57.8  & 54.9  & 43.5  & 33.9 \\
    & 5-shot  & 61.6   & 59.4  & 42.3  & 37.1  & 71.2  & 65.1  & 46.2  & 40.9 \\
    & 10-shot & 66.0   & 64.9  & 44.1  & 39.8  & 68.2  & 67.4  & 49.4  & 44.9 \\
\midrule
   \multirow{3}{*}{\textbf{\textsc{Tolar-QE}} (GPT) }
    & 1-shot  & 64.8   & 58.7  & 38.4  & 31.3  & 62.8  & 61.2  & 43.8  & 36.1  \\
    & 5-shot  & 67.5   & 59.6  & 41.4  & 36.5  & 66.1  & 66.1  & 47.5  & 43.6  \\
    & 10-shot & 67.4   & 65.2  & 42.7  & 39.1  & 71.1  & 70.4  & 49.2  & 46.5  \\
\bottomrule
\end{tabular}
\caption{Few-shot Event Extraction results with data augmentation on ACE05-E with QueryExtract (QE).}
\label{tab:fs_ace_qe}
\end{table*}

\section{Negative Event Mentions Prompts}
\label{appendix:neg_event_mentions}
Table \ref{tab:negative_prompt} list generation instructions of negative event mentions for generation agents. Table \ref{tab:neg_aug_example} shows negative augmentation examples.

\section{Experimental Results with QE}
Table \ref{tab:fs_ace_qe} shows Experimental results for ACE05-E with QueryExtract (QE) as the baseline model. 

\section{Features Contributed by Augmented Data}
{The features that are better captured by the proposed approach include (1) The mapping between candidate triggers and event types. The presence of a greater variety of event mention expressions within diverse contexts enhances the robustness and comprehensiveness of the mapping between candidate triggers and event types. (2) The mapping between negative expressions and event types. Due to the limited occurrence of negative events in the training data, their availability as few-shot examples is restricted. With the integration of the negative augmentation module, the mapping between negative expressions and event types becomes clearer. (3) The relation between candidate triggers and arguments. The generated sentences exhibit a comparatively higher prevalence of straightforward event expressions than those present in annotated data, such as ACE2005. These less complex expressions contribute to a good fit for features related to the relation between candidate triggers and arguments, in the low-resource settings.}

%% file: tables/neg_aug_example.tex
\begin{table*}[ht!]
\begin{center}
\small
\begin{tabular}{c|lP}
\toprule
id&\textbf{Note}&\textbf{Content} \\
\midrule
\multirow{9}{*}{1}
&Event Structure& Trigger: bankruptcy. Org: Hazelhurst \& Associates Inc.\\
&Context& 10 percent, yesterday, \$22.5 million
\\\cmidrule{2-3}
&Positive mention&
Hazelhurst \& Associates Inc. declared bankruptcy yesterday, with \$22.5 million in debts.
\\
&Negative mention&
Hazelhurst \& Associates Inc. did not declare bankruptcy yesterday, with \$22.5 million in debts.
\\
&Asserted mention:&
It is believed that Hazelhurst \& Associates Inc. will declare bankruptcy tomorrow, with \$30 million in debts.
\\
\midrule
\multirow{9}{*}{2}&Event Structure& Trigger: pardon, Place: Jordan, Adjudicator: Abdullah II,  Defendant: Rich\\
&Context& Republicans, today, his darkest hours\\\cmidrule{2-3}
&Positive mention& Rich received a pardon from Abdullah II during his darkest hours , as Republicans gathered today to offer their support . \\
&Negative mention& Rich's pardon from Abdullah II was canceled during his darkest hours, as Republicans did not gather.\\
&Asserted mention& Rich desired to receive a pardon from Abdullah II during his darkest hours, as Republicans gathered last year to offer their support.\\
\bottomrule
\end{tabular}

\caption{Negative Augmentation Example}
\label{tab:neg_aug_example}
\end{center}
\end{table*}

%% file: main.bbl
\begin{thebibliography}{70}
\expandafter\ifx\csname natexlab\endcsname\relax\def\natexlab#1{#1}\fi

\bibitem[{Agarwal et~al.(2021)Agarwal, Majee, Subramanian, and Arora}]{agarwal2021attention}
Ashutosh Agarwal, Anay Majee, Anbumani Subramanian, and Chetan Arora. 2021.
\newblock \href {http://arxiv.org/abs/2111.06639} {Attention guided cosine margin for overcoming class-imbalance in few-shot road object detection}.

\bibitem[{Ahn(2006)}]{ahn2006stages}
David Ahn. 2006.
\newblock The stages of event extraction.
\newblock In \emph{Proceedings of the Workshop on Annotating and Reasoning about Time and Events}, pages 1--8.

\bibitem[{Cao et~al.(2015)Cao, Li, Fan, and Grishman}]{cao-etal-2015-improving}
Kai Cao, Xiang Li, Miao Fan, and Ralph Grishman. 2015.
\newblock \href {https://aclanthology.org/R15-1010} {Improving event detection with active learning}.
\newblock In \emph{Proceedings of the International Conference Recent Advances in Natural Language Processing}, pages 72--77, Hissar, Bulgaria. INCOMA Ltd. Shoumen, BULGARIA.

\bibitem[{Cao and Wang(2021)}]{Cao_2021_acl_controllable}
Shuyang Cao and Lu~Wang. 2021.
\newblock \href {https://doi.org/10.18653/v1/2021.acl-long.502} {Controllable open-ended question generation with a new question type ontology}.

\bibitem[{Chen et~al.(2021)Chen, Lin, Han, and Sun}]{honey_chen_2021}
Jiawei Chen, Hongyu Lin, Xianpei Han, and Le~Sun. 2021.
\newblock Honey or poison? solving the trigger curse in few-shot event detection via causal intervention.

\bibitem[{Chen et~al.(2020)Chen, Chen, Ebner, White, and Van~Durme}]{chen-etal-2020-reading}
Yunmo Chen, Tongfei Chen, Seth Ebner, Aaron~Steven White, and Benjamin Van~Durme. 2020.
\newblock \href {https://doi.org/10.18653/v1/2020.spnlp-1.9} {Reading the manual: Event extraction as definition comprehension}.
\newblock In \emph{Proceedings of the Fourth Workshop on Structured Prediction for NLP}, Online. Association for Computational Linguistics.

\bibitem[{Cheng et~al.(2021)Cheng, Zhu, Li, Gong, Sun, and Liu}]{Cheng2020LearningWI}
Hao Cheng, Zhaowei Zhu, Xingyu Li, Yifei Gong, Xing Sun, and Yang Liu. 2021.
\newblock Learning with instance-dependent label noise: A sample sieve approach.
\newblock In \emph{International Conference on Learning Representations}.

\bibitem[{Chinchor and Marsh(1998)}]{chinchor1998muc}
Nancy Chinchor and Elaine Marsh. 1998.
\newblock Muc-7 information extraction task definition.
\newblock In \emph{Proceeding of the seventh message understanding conference (MUC-7), Appendices}, pages 359--367.

\bibitem[{Chowdhury et~al.(2021)Chowdhury, Jiang, and Jermaine}]{Chowdhury2021FewshotIC}
Arkabandhu Chowdhury, Mingchao Jiang, and Chris Jermaine. 2021.
\newblock Few-shot image classification: Just use a library of pre-trained feature extractors and a simple classifier.

\bibitem[{Cong et~al.(2021)Cong, Cui, Yu, Liu, Wang, and Wang}]{Cong2021FewShotED}
Xin Cong, Shiyao Cui, Bowen Yu, Tingwen Liu, Yubin Wang, and Bin Wang. 2021.
\newblock Few-shot event detection with prototypical amortized conditional random field.
\newblock In \emph{Findings of the Association for Computational Linguistics: ACL-IJCNLP}.

\bibitem[{Deng et~al.(2020)Deng, Zhang, Kang, Zhang, Zhang, and Chen}]{Meta_Deng_2020}
Shumin Deng, Ningyu Zhang, Jiaojian Kang, Yichi Zhang, Wei Zhang, and Huajun Chen. 2020.
\newblock \href {https://doi.org/10.1145/3336191.3371796} {Meta-learning with dynamic-memory-based prototypical network for few-shot event detection}.
\newblock \emph{Proceedings of the 13th International Conference on Web Search and Data Mining}.

\bibitem[{Du and Cardie(2020)}]{xinyaduEMNLP2020}
Xinya Du and Claire Cardie. 2020.
\newblock \href {https://doi.org/10.18653/v1/2020.emnlp-main.49} {Event extraction by answering (almost) natural questions}.
\newblock In \emph{Proceedings of the 2020 Conference on Empirical Methods in Natural Language Processing (EMNLP)}, pages 671--683, Online. Association for Computational Linguistics.

\bibitem[{Edunov et~al.(2018)Edunov, Ott, Auli, and Grangier}]{edunov-etal-2018-understanding}
Sergey Edunov, Myle Ott, Michael Auli, and David Grangier. 2018.
\newblock \href {https://doi.org/10.18653/v1/D18-1045} {Understanding back-translation at scale}.
\newblock In \emph{Proceedings of the 2018 Conference on Empirical Methods in Natural Language Processing}, pages 489--500, Brussels, Belgium. Association for Computational Linguistics.

\bibitem[{Edwards et~al.(2021)Edwards, Ushio, Camacho-Collados, de~Ribaupierre, and Preece}]{Edwards_2021_Guiding}
Aleksandra Edwards, Asahi Ushio, Jose Camacho-Collados, Hélène de~Ribaupierre, and Alun Preece. 2021.
\newblock \href {https://doi.org/10.48550/ARXIV.2111.09064} {Guiding generative language models for data augmentation in few-shot text classification}.

\bibitem[{Feng et~al.(2020)Feng, Gangal, Kang, Mitamura, and Hovy}]{feng-etal-2020-genaug}
Steven~Y. Feng, Varun Gangal, Dongyeop Kang, Teruko Mitamura, and Eduard Hovy. 2020.
\newblock \href {https://doi.org/10.18653/v1/2020.deelio-1.4} {{G}en{A}ug: Data augmentation for finetuning text generators}.
\newblock In \emph{Proceedings of Deep Learning Inside Out (DeeLIO): The First Workshop on Knowledge Extraction and Integration for Deep Learning Architectures}, pages 29--42, Online. Association for Computational Linguistics.

\bibitem[{Gao et~al.(2022)Gao, Sagawa, Koh, Hashimoto, and Liang}]{gao2022outofdistribution}
Irena Gao, Shiori Sagawa, Pang~Wei Koh, Tatsunori Hashimoto, and Percy Liang. 2022.
\newblock \href {https://openreview.net/forum?id=Bcg0It4i1g} {Out-of-distribution robustness via targeted augmentations}.
\newblock In \emph{NeurIPS 2022 Workshop on Distribution Shifts: Connecting Methods and Applications}.

\bibitem[{Ghosh et~al.(2021)Ghosh, Qi, Chaturvedi, and Srivastava}]{ghosh-etal-2021-helpful}
Sayan Ghosh, Zheng Qi, Snigdha Chaturvedi, and Shashank Srivastava. 2021.
\newblock \href {https://doi.org/10.18653/v1/2021.acl-short.11} {How helpful is inverse reinforcement learning for table-to-text generation?}
\newblock In \emph{Proceedings of the 59th Annual Meeting of the Association for Computational Linguistics and the 11th International Joint Conference on Natural Language Processing (Volume 2: Short Papers)}, pages 71--79, Online. Association for Computational Linguistics.

\bibitem[{Gowda and May(2020)}]{gowda-may-2020-finding}
Thamme Gowda and Jonathan May. 2020.
\newblock \href {https://doi.org/10.18653/v1/2020.findings-emnlp.352} {Finding the optimal vocabulary size for neural machine translation}.
\newblock In \emph{Findings of the Association for Computational Linguistics: EMNLP 2020}, pages 3955--3964, Online. Association for Computational Linguistics.

\bibitem[{Grishman(1997)}]{grishman1997information}
Ralph Grishman. 1997.
\newblock Information extraction: Techniques and challenges.
\newblock In \emph{International summer school on information extraction}, pages 10--27. Springer.

\bibitem[{Guan et~al.(2021)Guan, Mao, Fan, Liu, Ding, and Huang}]{Guan2021LongTG}
Jian Guan, Xiao-Xi Mao, Changjie Fan, Zitao Liu, Wenbiao Ding, and Minlie Huang. 2021.
\newblock Long text generation by modeling sentence-level and discourse-level coherence.
\newblock In \emph{ACL}.

\bibitem[{Han et~al.(2018)Han, Yao, Yu, Niu, Xu, Hu, Tsang, and Sugiyama}]{Han_coteaching}
Bo~Han, Quanming Yao, Xingrui Yu, Gang Niu, Miao Xu, Weihua Hu, Ivor~W. Tsang, and Masashi Sugiyama. 2018.
\newblock Co-teaching: Robust training of deep neural networks with extremely noisy labels.
\newblock In \emph{Proceedings of the 32nd International Conference on Neural Information Processing Systems}, NIPS'18, page 8536–8546, Red Hook, NY, USA. Curran Associates Inc.

\bibitem[{Han et~al.(2023)Han, Peng, Yang, Wang, Liu, and Wan}]{han2023information}
Ridong Han, Tao Peng, Chaohao Yang, Benyou Wang, Lu~Liu, and Xiang Wan. 2023.
\newblock \href {http://arxiv.org/abs/2305.14450} {Is information extraction solved by chatgpt? an analysis of performance, evaluation criteria, robustness and errors}.

\bibitem[{Hsu et~al.(2022)Hsu, Huang, Boschee, Miller, Natarajan, Chang, and Peng}]{naacl2022degree}
I-Hung Hsu, Kuan-Hao Huang, Elizabeth Boschee, Scott Miller, Prem Natarajan, Kai-Wei Chang, and Nanyun Peng. 2022.
\newblock Degree: A data-efficient generative event extraction model.
\newblock In \emph{Proceedings of the 2022 Conference of the North American Chapter of the Association for Computational Linguistics (NAACL)}.

\bibitem[{Hu et~al.(2021{\natexlab{a}})Hu, Zhang, Ma, Liu, Wen, and Yu}]{Hu-etal-2021-semi}
Xuming Hu, Chenwei Zhang, Fukun Ma, Chenyao Liu, Lijie Wen, and Philip~S. Yu. 2021{\natexlab{a}}.
\newblock Semi-supervised relation extraction via incremental meta self-training.
\newblock In \emph{Findings of the Association for Computational Linguistics: EMNLP 2021}, Online and Punta Cana, Dominican Republic. Association for Computational Linguistics.

\bibitem[{Hu et~al.(2021{\natexlab{b}})Hu, Zhang, Yang, Li, Lin, Wen, and Yu}]{Hu-etal-2021-gradient}
Xuming Hu, Chenwei Zhang, Yawen Yang, Xiaohe Li, Li~Lin, Lijie Wen, and Philip~S. Yu. 2021{\natexlab{b}}.
\newblock Gradient imitation reinforcement learning for low resource relation extraction.
\newblock In \emph{Proceedings of the 2021 Conference on Empirical Methods in Natural Language Processing}, Online and Punta Cana, Dominican Republic. Association for Computational Linguistics.

\bibitem[{Huang et~al.(2019)Huang, Qu, Jia, and Zhao}]{9008796}
Jinchi Huang, Lie Qu, Rongfei Jia, and Binqiang Zhao. 2019.
\newblock \href {https://doi.org/10.1109/ICCV.2019.00342} {O2u-net: A simple noisy label detection approach for deep neural networks}.
\newblock In \emph{2019 IEEE/CVF International Conference on Computer Vision (ICCV)}, pages 3325--3333.

\bibitem[{Huang et~al.(2016)Huang, Cassidy, Feng, Ji, Voss, Han, and Sil}]{huang2016liberal}
Lifu Huang, Taylor Cassidy, Xiaocheng Feng, Heng Ji, Clare Voss, Jiawei Han, and Avirup Sil. 2016.
\newblock Liberal event extraction and event schema induction.
\newblock In \emph{Proceedings of the 54th Annual Meeting of the Association for Computational Linguistics (Volume 1: Long Papers)}, pages 258--268.

\bibitem[{Huang and Ji(2020)}]{huang2020semi}
Lifu Huang and Heng Ji. 2020.
\newblock Semi-supervised new event type induction and event detection.
\newblock In \emph{Proceedings of the 2020 Conference on Empirical Methods in Natural Language Processing (EMNLP)}, pages 718--724.

\bibitem[{Hussein et~al.(2022)Hussein, Chowdhury, Abdelali, Dehak, and Ali}]{Hussein_2022_codeswitching}
Amir Hussein, Shammur~Absar Chowdhury, Ahmed Abdelali, Najim Dehak, and Ahmed Ali. 2022.
\newblock \href {https://doi.org/10.48550/ARXIV.2201.02550} {Code-switching text augmentation for multilingual speech processing}.

\bibitem[{Jiang et~al.(2020)Jiang, Huang, Liu, and Yang}]{jiang2020beyond}
Lu~Jiang, Di~Huang, Mason Liu, and Weilong Yang. 2020.
\newblock Beyond synthetic noise: Deep learning on controlled noisy labels.
\newblock In \emph{ICML}.

\bibitem[{Kang et~al.(2019)Kang, Liu, Wang, Yu, Feng, and Darrell}]{Kang_ICCV_2019}
Bingyi Kang, Zhuang Liu, Xin Wang, Fisher Yu, Jiashi Feng, and Trevor Darrell. 2019.
\newblock \href {https://doi.org/10.1109/ICCV.2019.00851} {Few-shot object detection via feature reweighting}.
\newblock In \emph{2019 IEEE/CVF International Conference on Computer Vision (ICCV)}, pages 8419--8428.

\bibitem[{Kim et~al.(2022)Kim, Woo, Oh, Cha, and Han}]{alp_aaai2022}
Hazel Kim, Daecheol Woo, Seong~Joon Oh, Jeong-Won Cha, and Yo-Sub Han. 2022.
\newblock \href {https://doi.org/10.1609/aaai.v36i10.21336} {Alp: Data augmentation using lexicalized pcfgs for few-shot text classification}.
\newblock \emph{Proceedings of the AAAI Conference on Artificial Intelligence}, 36:10894--10902.

\bibitem[{Lai et~al.(2020{\natexlab{a}})Lai, Dernoncourt, and Nguyen}]{Exploiting_Lai_2020}
Viet~Dac Lai, Franck Dernoncourt, and Thien~Huu Nguyen. 2020{\natexlab{a}}.
\newblock \href {https://doi.org/10.1007/978-3-030-47436-2_18} {Exploiting the matching information in the support set for few shot event classification}.
\newblock \emph{Pacific-Asia Conference on Knowledge Discovery and Data Mining}, page 233–245.

\bibitem[{Lai et~al.(2020{\natexlab{b}})Lai, Nguyen, and Dernoncourt}]{extensively_lai_2020}
Viet~Dac Lai, Thien~Huu Nguyen, and Franck Dernoncourt. 2020{\natexlab{b}}.
\newblock \href {https://doi.org/10.18653/v1/2020.nuse-1.5} {Extensively matching for few-shot learning event detection}.
\newblock In \emph{Proceedings of the First Joint Workshop on Narrative Understanding, Storylines, and Events}, pages 38--45, Online. Association for Computational Linguistics.

\bibitem[{Li et~al.(2023)Li, Fang, Yang, Wang, Ye, Zhao, and Zhang}]{li2023evaluating}
Bo~Li, Gexiang Fang, Yang Yang, Quansen Wang, Wei Ye, Wen Zhao, and Shikun Zhang. 2023.
\newblock \href {http://arxiv.org/abs/2304.11633} {Evaluating chatgpt's information extraction capabilities: An assessment of performance, explainability, calibration, and faithfulness}.

\bibitem[{Li et~al.(2021)Li, Yang, Liu, Liu, Ji, and Ye}]{Li2021BeyondMC}
Bohao Li, Boyu Yang, Chang Liu, Feng Liu, Rongrong Ji, and Qixiang Ye. 2021.
\newblock Beyond max-margin: Class margin equilibrium for few-shot object detection.
\newblock \emph{2021 IEEE/CVF Conference on Computer Vision and Pattern Recognition (CVPR)}, pages 7359--7368.

\bibitem[{Li et~al.(2020{\natexlab{a}})Li, Peng, Chen, Wang, Pan, Lyu, and Zhu}]{EEMQA_li}
Fayuan Li, Weihua Peng, Yuguang Chen, Quan Wang, Lu~Pan, Yajuan Lyu, and Yong Zhu. 2020{\natexlab{a}}.
\newblock \href {https://doi.org/10.18653/v1/2020.findings-emnlp.73} {Event extraction as multi-turn question answering}.
\newblock In \emph{Findings of the Association for Computational Linguistics: EMNLP 2020}, pages 829--838, Online. Association for Computational Linguistics.

\bibitem[{Li et~al.(2020{\natexlab{b}})Li, Socher, and Hoi}]{Li2020DivideMix}
Junnan Li, Richard Socher, and Steven~C.H. Hoi. 2020{\natexlab{b}}.
\newblock \href {https://openreview.net/forum?id=HJgExaVtwr} {Dividemix: Learning with noisy labels as semi-supervised learning}.
\newblock In \emph{International Conference on Learning Representations}.

\bibitem[{Lin et~al.(2020)Lin, Ji, Huang, and Wu}]{yinglinACL2020}
Ying Lin, Heng Ji, Fei Huang, and Lingfei Wu. 2020.
\newblock \href {https://doi.org/10.18653/v1/2020.acl-main.713} {A joint neural model for information extraction with global features}.
\newblock In \emph{Proceedings of the 58th Annual Meeting of the Association for Computational Linguistics}, pages 7999--8009, Online. Association for Computational Linguistics.

\bibitem[{{Linguistic Data Consortium}(2005)}]{ldc_ace05}
{Linguistic Data Consortium}. 2005.
\newblock English annotation guidelines for events.
\newblock \url{https://www.ldc.upenn.edu/ sites/www.ldc.upenn.edu/files/ english-events-guidelines-v5.4.3. pdf.}

\bibitem[{Liu et~al.(2020)Liu, Chen, Liu, Bi, and Liu}]{jianliu2020emnlp}
Jian Liu, Yubo Chen, Kang Liu, Wei Bi, and Xiaojiang Liu. 2020.
\newblock \href {https://doi.org/10.18653/v1/2020.emnlp-main.128} {Event extraction as machine reading comprehension}.
\newblock In \emph{Proceedings of the 2020 Conference on Empirical Methods in Natural Language Processing (EMNLP)}, pages 1641--1651, Online. Association for Computational Linguistics.

\bibitem[{Liu et~al.(2022)Liu, Huang, Shi, and Wang}]{liu-etal-2022-dynamic}
Xiao Liu, Heyan Huang, Ge~Shi, and Bo~Wang. 2022.
\newblock \href {https://doi.org/10.18653/v1/2022.acl-long.358} {Dynamic prefix-tuning for generative template-based event extraction}.
\newblock In \emph{Proceedings of the 60th Annual Meeting of the Association for Computational Linguistics (Volume 1: Long Papers)}, pages 5216--5228, Dublin, Ireland. Association for Computational Linguistics.

\bibitem[{Loem et~al.(2022)Loem, Takase, Kaneko, and Okazaki}]{Loem_2022_ExtraPhrase}
Mengsay Loem, Sho Takase, Masahiro Kaneko, and Naoaki Okazaki. 2022.
\newblock \href {https://doi.org/10.48550/ARXIV.2201.05313} {Extraphrase: Efficient data augmentation for abstractive summarization}.

\bibitem[{Lyu et~al.(2021)Lyu, Zhang, Sulem, and Roth}]{Lyu-etal-2021-zero}
Qing Lyu, Hongming Zhang, Elior Sulem, and Dan Roth. 2021.
\newblock {Z}ero-shot {E}vent {E}xtraction via {T}ransfer {L}earning: {C}hallenges and {I}nsights.
\newblock In \emph{Proceedings of the 59th Annual Meeting of the Association for Computational Linguistics and the 11th International Joint Conference on Natural Language Processing (Volume 2: Short Papers)}, pages 322--332, Online. Association for Computational Linguistics.

\bibitem[{Ma et~al.(2023)Ma, Wang, Kung, Brantingham, Peng, and Wang}]{ma2023star}
Mingyu~Derek Ma, Xiaoxuan Wang, Po-Nien Kung, P.~Jeffrey Brantingham, Nanyun Peng, and Wei Wang. 2023.
\newblock \href {http://arxiv.org/abs/2305.15090} {Star: Improving low-resource information extraction by structure-to-text data generation with large language models}.

\bibitem[{Ng et~al.(2020)Ng, Cho, and Ghassemi}]{ng-etal-2020-ssmba}
Nathan Ng, Kyunghyun Cho, and Marzyeh Ghassemi. 2020.
\newblock \href {https://doi.org/10.18653/v1/2020.emnlp-main.97} {{SSMBA}: Self-supervised manifold based data augmentation for improving out-of-domain robustness}.
\newblock In \emph{Proceedings of the 2020 Conference on Empirical Methods in Natural Language Processing (EMNLP)}, pages 1268--1283, Online. Association for Computational Linguistics.

\bibitem[{Pasupat and Liang(2014)}]{pasupat2014zero}
Panupong Pasupat and Percy Liang. 2014.
\newblock Zero-shot entity extraction from web pages.
\newblock In \emph{Proceedings of the 52nd Annual Meeting of the Association for Computational Linguistics (Volume 1: Long Papers)}, pages 391--401.

\bibitem[{Shen et~al.(2021{\natexlab{a}})Shen, Meng, Zhang, Feng, and Zhou}]{Shen2021GTMAG}
Lei Shen, Fandong Meng, Jinchao Zhang, Yang Feng, and Jie Zhou. 2021{\natexlab{a}}.
\newblock \href {https://doi.org/10.18653/v1/2021.acl-long.271} {{GTM}: A generative triple-wise model for conversational question generation}.
\newblock In \emph{Proceedings of the 59th Annual Meeting of the Association for Computational Linguistics and the 11th International Joint Conference on Natural Language Processing (Volume 1: Long Papers)}, pages 3495--3506, Online. Association for Computational Linguistics.

\bibitem[{Shen et~al.(2021{\natexlab{b}})Shen, Wu, Qi, Li, Haffari, and Bi}]{Adaptive_Shirong_2021}
Shirong Shen, Tongtong Wu, Guilin Qi, Yuan-Fang Li, Gholamreza Haffari, and Sheng Bi. 2021{\natexlab{b}}.
\newblock \href {https://doi.org/10.18653/v1/2021.findings-acl.214} {Adaptive knowledge-enhanced bayesian meta-learning for few-shot event detection}.
\newblock In \emph{Findings of the Association for Computational Linguistics}, page 2417–2429. Association for Computational Linguistics (ACL).

\bibitem[{Song et~al.(2015)Song, Bies, Strassel, Riese, Mott, Ellis, Wright, Kulick, Ryant, and Ma}]{song2015light}
Zhiyi Song, Ann Bies, Stephanie Strassel, Tom Riese, Justin Mott, Joe Ellis, Jonathan Wright, Seth Kulick, Neville Ryant, and Xiaoyi Ma. 2015.
\newblock From light to rich ere: annotation of entities, relations, and events.
\newblock In \emph{Proceedings of the the 3rd Workshop on EVENTS: Definition, Detection, Coreference, and Representation}, pages 89--98.

\bibitem[{Sun et~al.(2021)Sun, Li, Cai, Yuan, and Zhang}]{Sun2021FSCEFO}
Bo~Sun, Banghuai Li, Shengcai Cai, Ye~Yuan, and Chi Zhang. 2021.
\newblock Fsce: Few-shot object detection via contrastive proposal encoding.
\newblock \emph{2021 IEEE/CVF Conference on Computer Vision and Pattern Recognition (CVPR)}, pages 7348--7358.

\bibitem[{Wang et~al.(2023{\natexlab{a}})Wang, Huang, Wei, Shi, Liu, Feng, Zhou, Wang, and Yin}]{wang-etal-2023-boosting}
Bo~Wang, Heyan Huang, Xiaochi Wei, Ge~Shi, Xiao Liu, Chong Feng, Tong Zhou, Shuaiqiang Wang, and Dawei Yin. 2023{\natexlab{a}}.
\newblock \href {https://doi.org/10.18653/v1/2023.findings-acl.716} {Boosting event extraction with denoised structure-to-text augmentation}.
\newblock In \emph{Findings of the Association for Computational Linguistics: ACL 2023}, pages 11267--11281, Toronto, Canada. Association for Computational Linguistics.

\bibitem[{Wang et~al.(2022)Wang, Yu, Chang, Sun, and Huang}]{WangAcl2022_query}
Sijia Wang, Mo~Yu, Shiyu Chang, Lichao Sun, and Lifu Huang. 2022.
\newblock \href {https://doi.org/10.18653/v1/2022.findings-acl.16} {Query and extract: Refining event extraction as type-oriented binary decoding}.
\newblock In \emph{Findings of the Association for Computational Linguistics: ACL 2022}, pages 169--182, Dublin, Ireland. Association for Computational Linguistics.

\bibitem[{Wang et~al.(2023{\natexlab{b}})Wang, Yu, and Huang}]{wang-etal-2023-art}
Sijia Wang, Mo~Yu, and Lifu Huang. 2023{\natexlab{b}}.
\newblock \href {https://doi.org/10.18653/v1/2023.acl-short.111} {The art of prompting: Event detection based on type specific prompts}.
\newblock In \emph{Proceedings of the 61st Annual Meeting of the Association for Computational Linguistics (Volume 2: Short Papers)}, pages 1286--1299, Toronto, Canada. Association for Computational Linguistics.

\bibitem[{Wang et~al.(2020{\natexlab{a}})Wang, Huang, Darrell, Gonzalez, and Yu}]{wang2020few}
Xin Wang, Thomas~E. Huang, Trevor Darrell, Joseph~E Gonzalez, and Fisher Yu. 2020{\natexlab{a}}.
\newblock Frustratingly simple few-shot object detection.

\bibitem[{Wang et~al.(2021{\natexlab{a}})Wang, Zheng, Jiang, and Huang}]{wang2021adalabel}
Yida Wang, Yinhe Zheng, Yong Jiang, and Minlie Huang. 2021{\natexlab{a}}.
\newblock Diversifying dialog generation via adaptive label smoothing.
\newblock In \emph{Proceedings of the 59th Annual Meeting of the Association for Computational Linguistics}.

\bibitem[{Wang et~al.(2021{\natexlab{b}})Wang, Wood, Wan, Dras, and Johnson}]{MentionFlags}
Yufei Wang, {Ian D.} Wood, Stephen Wan, Mark Dras, and Mark Johnson. 2021{\natexlab{b}}.
\newblock \href {https://doi.org/10.18653/v1/2021.acl-long.9} {Mention flags (mf): constraining transformer-based text generators}.
\newblock In \emph{Proceedings of the 59th Annual Meeting of the Association for Computational Linguistics and the 11th International Joint Conference on Natural Language Processing (Volume 1: Long Papers)}, ACL-IJCNLP 2021 - 59th Annual Meeting of the Association for Computational Linguistics and the 11th International Joint Conference on Natural Language Processing, Proceedings of the Conference, pages 103--113. Association for Computational Linguistics (ACL).

\bibitem[{Wang et~al.(2020{\natexlab{b}})Wang, Jiang, Han, Feng, An, Niu, and Long}]{Wang2020SemiNLLAF}
Zhuowei Wang, Jing Jiang, Bo~Han, Lei Feng, Bo~An, Gang Niu, and Guodong Long. 2020{\natexlab{b}}.
\newblock \href {https://api.semanticscholar.org/CorpusID:227248138} {Seminll: A framework of noisy-label learning by semi-supervised learning}.
\newblock \emph{Transactions on Machine Learning Research}, 2022.

\bibitem[{Wei et~al.(2020)Wei, Feng, Chen, and An}]{9156369}
Hongxin Wei, Lei Feng, Xiangyu Chen, and Bo~An. 2020.
\newblock \href {https://doi.org/10.1109/CVPR42600.2020.01374} {Combating noisy labels by agreement: A joint training method with co-regularization}.
\newblock In \emph{2020 IEEE/CVF Conference on Computer Vision and Pattern Recognition (CVPR)}, pages 13723--13732.

\bibitem[{Wei and Zou(2019)}]{wei-zou-2019-eda}
Jason Wei and Kai Zou. 2019.
\newblock {EDA}: Easy data augmentation techniques for boosting performance on text classification tasks.
\newblock In \emph{Proceedings of the 2019 Conference on Empirical Methods in Natural Language Processing and the 9th International Joint Conference on Natural Language Processing (EMNLP-IJCNLP)}, pages 6383--6389, Hong Kong, China. Association for Computational Linguistics.

\bibitem[{Wei et~al.(2023)Wei, Cui, Cheng, Wang, Zhang, Huang, Xie, Xu, Chen, Zhang, Jiang, and Han}]{wei2023zeroshot}
Xiang Wei, Xingyu Cui, Ning Cheng, Xiaobin Wang, Xin Zhang, Shen Huang, Pengjun Xie, Jinan Xu, Yufeng Chen, Meishan Zhang, Yong Jiang, and Wenjuan Han. 2023.
\newblock \href {http://arxiv.org/abs/2302.10205} {Zero-shot information extraction via chatting with chatgpt}.

\bibitem[{Xiao and Marlet(2020)}]{Xiao2020FewShotOD}
Yang Xiao and Renaud Marlet. 2020.
\newblock Few-shot object detection and viewpoint estimation for objects in the wild.
\newblock In \emph{ECCV}.

\bibitem[{Yan et~al.(2019)Yan, Chen, Xu, Wang, Liang, and Lin}]{Yan2019MetaRT}
Xiaopeng Yan, Ziliang Chen, Anni Xu, Xiaoxi Wang, Xiaodan Liang, and Liang Lin. 2019.
\newblock Meta r-cnn: Towards general solver for instance-level low-shot learning.
\newblock \emph{2019 IEEE/CVF International Conference on Computer Vision (ICCV)}, pages 9576--9585.

\bibitem[{Yang et~al.(2020)Yang, Malaviya, Fernandez, Swayamdipta, Le~Bras, Wang, Bhagavatula, Choi, and Downey}]{yang-etal-2020-generative}
Yiben Yang, Chaitanya Malaviya, Jared Fernandez, Swabha Swayamdipta, Ronan Le~Bras, Ji-Ping Wang, Chandra Bhagavatula, Yejin Choi, and Doug Downey. 2020.
\newblock \href {https://doi.org/10.18653/v1/2020.findings-emnlp.90} {Generative data augmentation for commonsense reasoning}.
\newblock In \emph{Findings of the Association for Computational Linguistics: EMNLP 2020}, pages 1008--1025, Online. Association for Computational Linguistics.

\bibitem[{Yao et~al.(2020)Yao, Yang, Han, Niu, and Kwok}]{Yao2020SearchingTE}
Quanming Yao, Hansi Yang, Bo~Han, Gang Niu, and James Tin-Yau Kwok. 2020.
\newblock \href {https://api.semanticscholar.org/CorpusID:221082209} {Searching to exploit memorization effect in learning with noisy labels}.
\newblock In \emph{International Conference on Machine Learning}.

\bibitem[{Yu et~al.(2019)Yu, Han, Yao, Niu, Tsang, and Sugiyama}]{yu2019does}
Xingrui Yu, Bo~Han, Jiangchao Yao, Gang Niu, Ivor Tsang, and Masashi Sugiyama. 2019.
\newblock How does disagreement help generalization against label corruption?
\newblock In \emph{International Conference on Machine Learning}, pages 7164--7173.

\bibitem[{Zhang et~al.(2021{\natexlab{a}})Zhang, Cui, Wu, Lu, and Tian}]{Zhang2021PNPDetEF}
Gongjie Zhang, Kaiwen Cui, Rongliang Wu, Shijian Lu, and Yonghong Tian. 2021{\natexlab{a}}.
\newblock Pnpdet: Efficient few-shot detection without forgetting via plug-and-play sub-networks.
\newblock \emph{2021 IEEE Winter Conference on Applications of Computer Vision (WACV)}, pages 3822--3831.

\bibitem[{Zhang et~al.(2021{\natexlab{b}})Zhang, Wang, and Roth}]{zhang-etal-2021-zero}
Hongming Zhang, Haoyu Wang, and Dan Roth. 2021{\natexlab{b}}.
\newblock \href {https://doi.org/10.18653/v1/2021.findings-acl.114} {{Z}ero-shot {L}abel-aware {E}vent {T}rigger and {A}rgument {C}lassification}.
\newblock In \emph{Findings of the Association for Computational Linguistics: ACL-IJCNLP 2021}, pages 1331--1340, Online. Association for Computational Linguistics.

\bibitem[{Zhang et~al.(2021{\natexlab{c}})Zhang, Lee, and Agarwal}]{Zhang2021LearningFN}
Mingyuan Zhang, Jane Lee, and Shivani Agarwal. 2021{\natexlab{c}}.
\newblock \href {https://api.semanticscholar.org/CorpusID:235826311} {Learning from noisy labels with no change to the training process}.
\newblock In \emph{International Conference on Machine Learning}.

\bibitem[{Zhu et~al.(2022)Zhu, Dong, and Liu}]{Zhu2022DetectingCL}
Zhaowei Zhu, Zihao Dong, and Yang Liu. 2022.
\newblock \href {https://api.semanticscholar.org/CorpusID:246431058} {Detecting corrupted labels without training a model to predict}.
\newblock In \emph{International Conference on Machine Learning}.

\end{thebibliography}
